\documentclass[lettersize,journal]{IEEEtran}
\usepackage{amsmath,amsfonts}
\usepackage{algorithmic}
\usepackage{algorithm}
\usepackage{array}
\usepackage[caption=false,font=normalsize,labelfont=sf,textfont=sf]{subfig}
\usepackage{textcomp}
\usepackage{stfloats}
\usepackage{url}
\usepackage{verbatim}
\usepackage{graphicx}
\usepackage{cite}
\usepackage{orcidlink}
\begin{document}

\title{Consensus-Driven Uncertainty for Robotic \\ Grasping based on RGB Perception}

\author{Eric C. Joyce$^{a}$ \orcidlink{0009-0000-4581-4399}, Qianwen Zhao$^{a}$ \orcidlink{0009-0003-0361-1181}, Nathaniel Burgdorfer$^{a}$ \orcidlink{0009-0004-4706-7373}, Long Wang$^{a}$ \orcidlink{0000-0003-3476-6779}, Philippos Mordohai$^{a}$ \orcidlink{0000-0002-9671-4408}%
\thanks{$^{a}$Stevens Institute of Technology, Hoboken, NJ 07030, USA.
       {\tt\small \{ejoyce, qzhao10, nburgdor, lwang4, pmordoha\}@stevens.edu}}%
\thanks{This research was supported in part by NSF Grant CMMI-2138896.}}

\maketitle

\begin{abstract}
Deep object pose estimators are notoriously overconfident. A grasping agent that both estimates the 6-DoF pose of a target object and predicts the uncertainty of its own estimate could avoid task failure by choosing not to act under high uncertainty. Even though object pose estimation improves and uncertainty quantification research continues to make strides, few studies have connected them to the downstream task of robotic grasping. We propose a method for training lightweight, deep networks to predict whether a grasp guided by an image-based pose estimate will succeed before that grasp is attempted. We generate training data for our networks via object pose estimation on real images and simulated grasping. We also find that, despite high object variability in grasping trials, networks benefit from training on all objects jointly, suggesting that a diverse variety of objects can nevertheless contribute to the same goal. Data, code, and guides are hosted at: \url{https://github.com/EricCJoyce/Consensus-Driven-Uncertainty/}
\end{abstract}

\section{Introduction}\label{sec:intro}
\begin{figure}[!t]
    \centering
    \includegraphics[width=0.9\columnwidth]{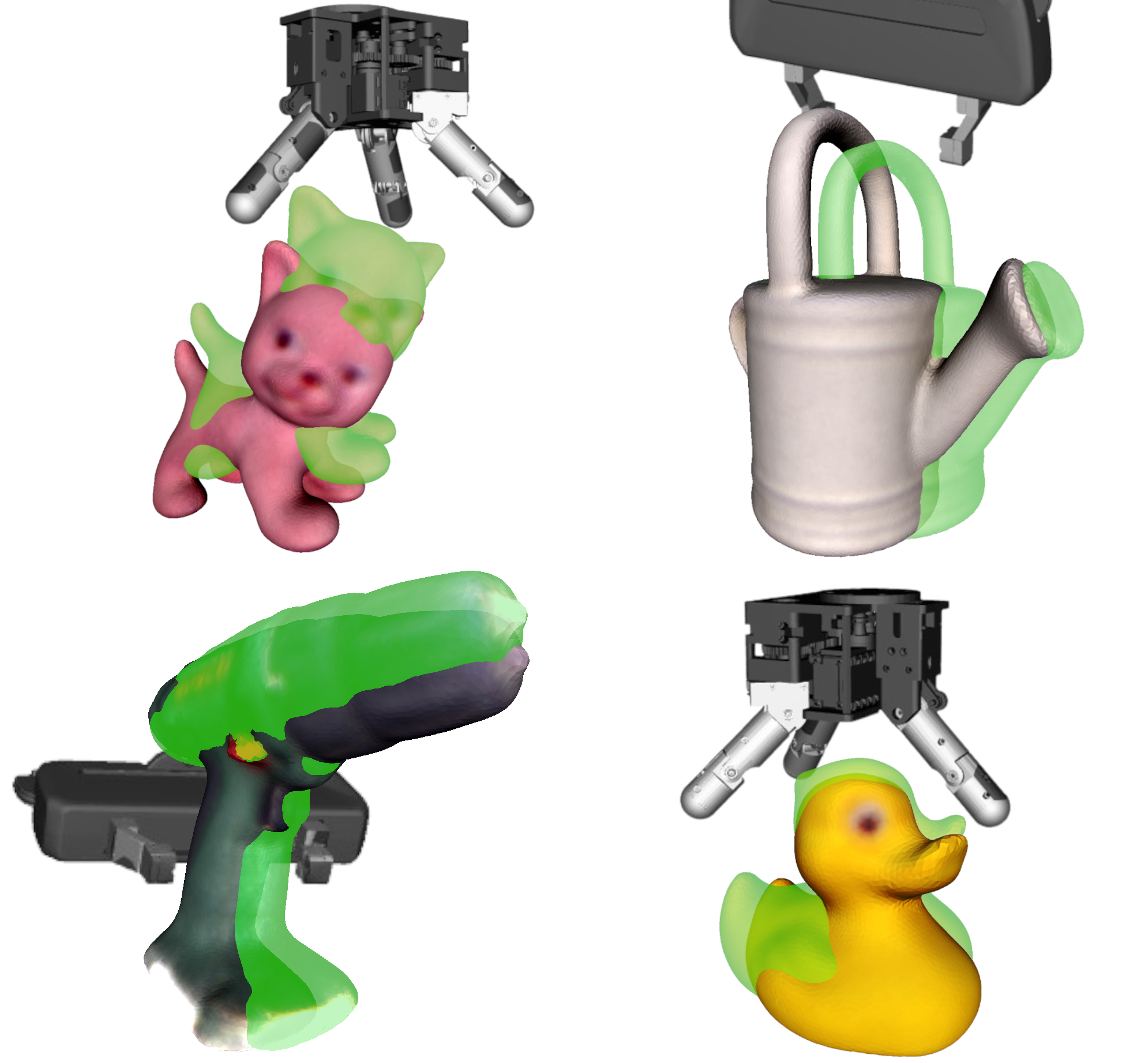}
    \vspace{-5pt}
    \caption{RGB-based pose estimates as green overlays and their corresponding ground-truth poses as solid objects. In our trials, grasping is attempted according to the estimated poses, and its success or failure is recorded as training data for 
    an uncertainty-quantification network. All estimates seen here perform better than average according to established metrics and yet contain errors sufficiently large to cause grasping trials to fail.}
    \vspace{-10pt}
    \label{fig:intro_figure_pose_errors}
\end{figure}

Remarkable progress in object pose estimation from single RGB images has been made in the past few years \cite{fan2022deep,  sahin2020review, sundermeyer2023bop, thalhammer2024challenges}, 
primarily driven by deep learning and the ability to reduce the so-called \textit{sim2real gap}. This has enabled end-to-end system training on large amounts of synthetic data with precise ground truth. Despite these advances and comprehensive benchmarks such as the Benchmark for 6D Object Pose Estimation (BOP) \cite{hodan2018bop, sundermeyer2023bop}\footnote{\url{https://bop.felk.cvut.cz/home/}}, grasping on the basis of these estimates is still difficult, and the potential for deploying RGB-based pose estimators in downstream robotic applications remains unclear. Consider for example the pose estimates illustrated in Figure~\ref{fig:intro_figure_pose_errors}. These were made by current methods, yet all four caused grasping attempts to fail when used as guides.

Motivated by this disconnect between pose evaluation and success in downstream grasping, we propose an approach to estimate the likelihood for success before a grasp is actually attempted. 
Our approach operates under the following assumptions: (i) perception occurs in RGB; (ii) objects are known and rigid; (iii) pose estimation occurs at the instance, not category, level; and (iv) a detector to locate the bounding box of the object in the image is available.
These settings are common in the robotics literature \cite{thalhammer2024challenges} and match those of BOP. 
Pose estimate accuracy naturally improves with the inclusion of a depth channel \cite{chao2021dexycb}, but RGB-only sensing is ubiquitous and applicable in both indoor and outdoor settings. 
A grasping agent equipped with a pose estimator aware of its own uncertainty would be able to avoid harmful failures by refusing to attempt grasps based on dubious estimates.

In their survey of under-studied challenges in monocular object pose estimation, Thalhammer et al. \cite{thalhammer2024challenges} urge researchers to focus on uncertainty in 6-DoF object pose estimation. One of the cited studies by Shi et al. \cite{shi2021fast} quantifies uncertainty but does not connect to the grasping task. In this paper, we propose a complete pipeline from perception, through uncertainty, to robotic grasping. This process begins by using off-the-shelf, high-performing 6-DoF pose estimators to make predictions about known objects in RGB images. As observed by the authors of FoundationPose \cite{wen2024foundationpose}, obtaining and accurately annotating real-world data is prohibitive. We therefore perform grasping trials in a physics simulator using the collected pose estimates, and assemble a training set for multi-layer perceptrons (MLP). These networks then learn to predict grasp failures. The result is conceptually similar to an ensemble but differs in that one of the ensemble members is promoted to the role of Principal Estimator (PE). We therefore refer to our approach as a ``consensus", wherein the PE is responsible for the downstream grasp, and the differences between the principal and other members' estimates becomes the basis for predicting the PE's success. 
Our approach is compatible with any number of object detectors, pose estimators, and grippers.
We compare our MLPs to a competitive baseline derived from Shi et al. and find that per-object grasp-failure prediction can be improved further by training one predictor jointly on all objects rather than individual, per-object predictors.

In short, our contributions are:
\begin{itemize}
    \item the integration of visual perception on real imagery, a grasping simulator, and uncertainty quantification, providing a complete pipeline for grasping agents to make and qualify predictions,
    \item a consensus-based type of predictor trained using off-the-shelf pose estimators and physics simulation results to assess a grasp attempt's probability for success,
    \item a grasping-trial dataset and means for collecting noise-free trial data.
\end{itemize}

\section{Related Work}\label{sec:related}
In this section, we focus on approaches that share the key characteristics of ours: RGB-only perception for pose estimation at the instance level, given known objects with an image detector, and a grasp per gripper for each object.
This puts adjacent research involving unknown objects 
\cite{lin2024sam,nguyen2024gigapose,wen2024foundationpose}, category-level pose estimation 
\cite{zhang2023genpose}, depth sensors 
\cite{deng2020self,zhang2023genpose,wen2024foundationpose}, grasp learning 
\cite{fang2020graspnet1b,sundermeyer2021contact,weng2023ngdf}, gripper design \cite{hagelskjaer2019combined}, and inverse kinematics and trajectory planning \cite{urain2023se3} out of scope.
Note that, unlike our method, some of these methods (e.g. FoundationPose) require a depth channel.

\textbf{RGB Pose Estimation Methods:}\label{sub::rgb-pose-est}
The progress made in pose estimation is evident in surveys \cite{fan2022deep, sahin2020review, thalhammer2024challenges} and on the BOP website. 
We categorize pose estimators into end-to-end (E2E) and PnP-based approaches. The former (e.g. \cite{xiang2017posecnn, sundermeyer2018implicit}) predict unmediated poses; the latter predict point correspondences which may be used to solve object poses using the Perspective-$n$-Points algorithm. E2E has been largely eclipsed by the PnP family.
PnP-based methods produce a number of correspondences sufficient to compute an object's pose.
``Sparse" estimators \cite{rad2017bb8, tremblay2018deep, tekin2018real} predict the projections of an object's 3D bounding-box corners or 3D control points. Pose is then deduced from these minimal enclosures.

When correspondences for almost every pixel are available, ``dense methods" such as \cite{peng2019pvnet, park2019pix2pose, hodan2020epos, wang2021gdr, su2022zebrapose, huang2022ncf, lian2023checkerpose} gain robustness to outliers. 
A dense PnP-based method, EPOS \cite{hodan2020epos} aims at robustness against textureless and symmetric objects by defining objects as sets of fragments. The network learns to predict probabilities for fragments to which a pixel might belong.
Geometry-Guided Direct Regression (GDR-Net) \cite{wang2021gdr} combines correspondence-based estimation and direct pose regression. GDR-Net generates dense 2D-3D correspondences as intermediate features before directly regressing pose using a learned, patch-based PnP approximator. Wang et al. credit their method's success to thoughtful representations for rotation \cite{zhou2019continuity} and translation \cite{li2019cdpn}, and to a loss function that combines pose and geometry.
ZebraPose \cite{su2022zebrapose} produces dense 2D-3D correspondences by first learning region-specific codes for object vertices. For all objects, vertices are partitioned into iteratively halved regions and assigned a binary code as a feature descriptor to be learned by an encoder-decoder. 
Once a network has been trained for each object, the decoder output for each pixel is the binary code of the 3D vertex corresponding to that 2D pixel. Pose is computed using these correspondences.
SurfEmb \cite{haugaard2022surfemb} adds a probabilistic aspect by learning to predict dense correspondence distributions over object surfaces. These distributions may then be sampled to form and refine pose hypotheses.
We use three of the above estimators in our experiments.

\textbf{Pose Uncertainty:}\label{sub::pose-uncertainty}
Pose uncertainty is often captured via sampling, as by Brachmann et al. \cite{brachmann2016uncertainty}, or modeled probabilistically, as in \cite{murphy2021implicitpdf, hofer2023hyperposepdf}. These studies express rotational ambiguity as multi-modal distributions over SO(3). How to also probabilistically bound a pose's translation (the SE(3) manifold) is less obvious. Haugaard et al. \cite{haugaard2023spyropose} propose a pyramid to model distributions over SE(3), constraining their distributions to the given object's radius. Our approach is deterministic and not limited to rotation uncertainty.

Ensembles \cite{lakshminarayanan2017ensembles} define uncertainty as a lack of consensus among networks trained for a common task. An ensemble-based uncertainty quantifier by Shi et al. \cite{shi2021fast} expresses pose uncertainty using the Averaged Distance between Distinguishable points (ADD). This approach, from which we construct a baseline method (see Section \ref{sub::baseline}), collapses the differences between ensemble members' estimates into a single value. Wursthorn et al. \cite{wursthorn2024uq} train an ensemble using SurfEmb \cite{haugaard2022surfemb} focusing on calibrated posterior probabilities.

\textbf{Downstream Grasping and Success Prediction:}\label{sub::downstream-grasping}
Pose uncertainty alone is not associated with any downstream task. The aforementioned methods quantify uncertainty, but there is no qualified pose estimate according to which an agent may choose to act.
Naik et al. \cite{naik2024robotic} model the likelihood of robotic grasping task success in the presence of pose-estimate error. They approach this probabilistically by perturbing the pose estimate and capturing how much error can be tolerated.

Though their method treats grasp optimization for unknown objects in RGB-D images (both out of scope), Marlier et al. \cite{marlier2024grasping} also predict probabilities for grasp success. They train a network to approximate the ratio of the likelihood and evidence, which are computationally intractable, and to find a grasp that improves the probability of success.

\textbf{Underactuated Robotic Hands:}\label{sub::robotic-hands-and-simulation}
In addition to the widely used Franka Hand \cite{haddadin2022franka}, a parallel gripper, we also simulate grasping trials with a tendon-driven underactuated hand 
\cite{Catalano2014,Ciocarlie2014,Dollar2010,Gosselin2008,Odhner2014,Stuart2017,wang2011highly}.
The mechanical compliance of these hands allows for a simplified, open-loop control scheme and adapts to object shape variations when grasping. The low cost and lightweight designs of underactuated hands enable use at scale. Compared to their counterpart, fully-actuated dexterous hands \cite{Jacobsen1986,Loucks1987,Deshpande2013}, underactuated hands can have higher and more realistic tolerance when object pose estimation errors are present. The underactuated hand in our work is a recent design \cite{chen2020underactuation}, the physics simulation of which has been used previously in deep reinforcement learning \cite{chen2020hardware}.

\section{Method}\label{sec:method}
\begin{figure*}[!h]
\centering
  \includegraphics[width=0.9\textwidth]{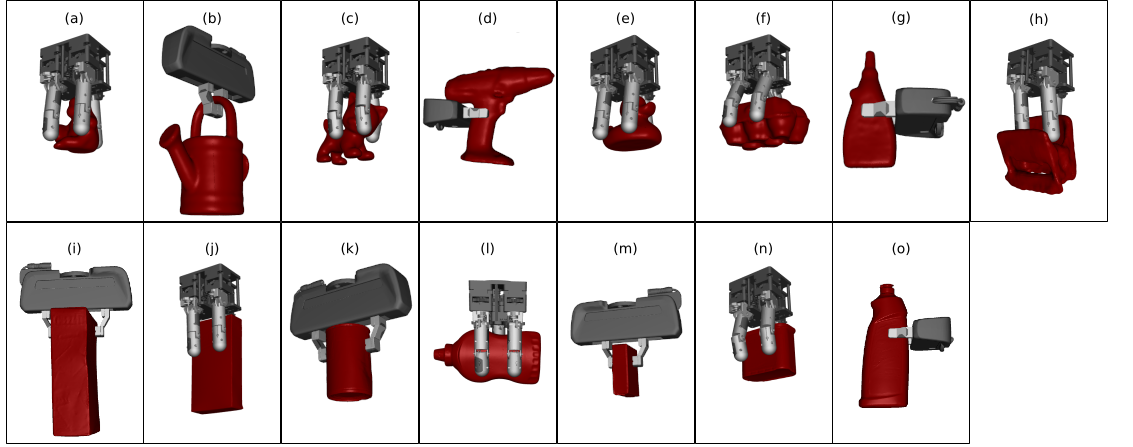}
  \caption{Example reference grasps for selected objects in the LM-O dataset (a-h) and YCB-V dataset (i-o). All grasping trials are attempted with both the parallel gripper, shown in (b), and the underactuated hand, shown in (a).}
  \vspace{-10pt}
  \label{fig:reference_grasps}
\end{figure*}

Our method trains multi-layer perceptrons (MLP) to predict whether a known gripper will succeed in grasping a known object that is visible in an input RGB image. Gathering the data necessary to train and test our networks begins with making pose estimates from RGB images and then applying those estimates to grasping trials in a physics simulator.

The proposed approach, once data have been collected, takes inspiration from the assumption guiding ensembles: that if several pose estimates differ only negligibly, then we may be sure about any one prediction. Conversely, if an ensemble's predictions vary substantially, then uncertainty about any single estimate exists. Our approach is not a true ensemble, however, because one estimator is elevated to the role of Principal. This is the estimator responsible for attempting a grasp if the probability for success is high enough.

\subsection{Object Pose Estimation}\label{sub::pose_est_from_rgb}

In the context of RGB images and known objects, grasping an object requires an estimate of its pose, which can be obtained by a 6-DoF pose estimator. Estimators receive real-world images of cluttered scenes containing instances of the target objects. The scene clutter and occlusions ensure that pose estimators operate under a variety of realistic conditions. Given a single image, an instance of a known, rigid object is detected in 2D. From the 2D bounding box of this instance, a pose estimator predicts a rotation and translation. Though detection is a necessary step in pose estimation, we consider an in-depth examination of detection out of scope for this paper. (The pose estimators in our experiments \cite{hodan2020epos,wang2021gdr,su2022zebrapose}, include their own detection results. Please see Section \ref{sub::estimators}.) Specifically, estimator $k$ yields a predicted pose $^{C_j}\hat{\mathbf{T}}_{O_i}^{(k)}$ for object $i$ as perceived in input image $j$. The ground-truth pose is $^{C_j}\mathbf{T}_{O_i}$. The notation signifies a rigid transformation from the frame of object $\{O_i\}$ to the frame of camera $\{C_j\}$.
The physics simulator receives each estimator's predicted pose and each corresponding ground-truth pose included in the dataset.

\begin{figure*}[!h]
\centering
  \includegraphics[width=0.95\textwidth]{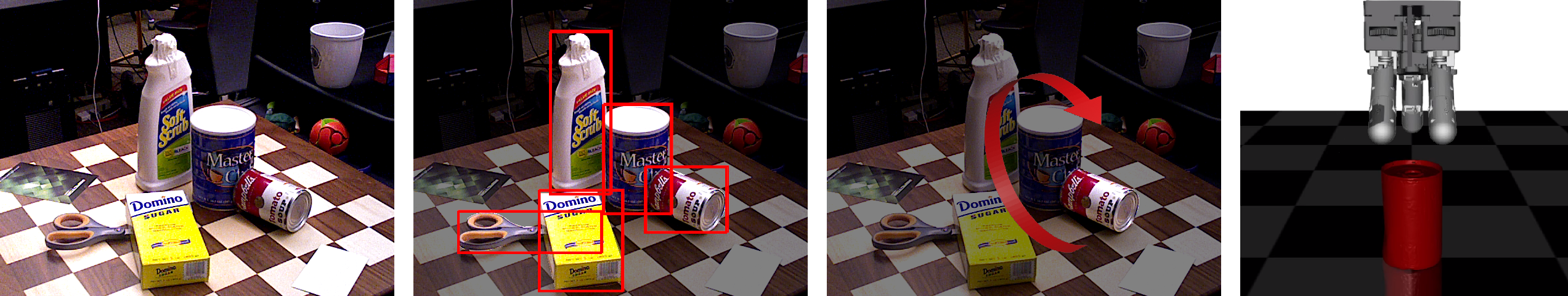}
  \caption{From RGB images, to object detections, to pose estimates per object, to grasping each object in a physics simulator. We abstract away scene clutter in order to isolate the effects of pose estimation on our downstream task. Given the reference grasp and pose we defined, this soup can should be upright for the grasping trial. Estimated and true poses are transformed so that the ground-truth pose aligns with a defined reference pose for that object, thus preserving pose estimates' deviations.}
  \vspace{-10pt}
  \label{fig:change-of-frame}
\end{figure*}

\subsection{Physics-based Grasping Simulation}\label{sub::simulation}

Because resetting precisely thousands of object poses in the real-world would introduce excessive noise, we conduct grasping trials in a physics simulator.
MuJoCo \cite{mujoco} is used to output a binary success indicator for each pose estimate $^{C_j}\hat{\mathbf{T}}_{O_i}^{(k)}$ and a given gripper $g$. 
The dataset includes 3D models of the objects but no information about their weights or friction coefficients. We have approximated values appropriate for each object and assumed that they have uniform density. 
The intuition is that we place an object in the simulation at its ground-truth pose, attempt to grasp it using $g$ according to the pose estimate made by $k$, and log the success or failure of the attempt. However, given that the true pose $^{C_j}\mathbf{T}_{O_i}$ is different for every frame $j$, this would require defining a target grasp for every $(i, j, g)$ in the dataset.
Furthermore, each separately defined grasp would need to be of comparable quality in order to single out the effects of pose-estimate error on the downstream task.

Rather than define as many grasps as there are trials, we equalize our trials by removing the influence of factors such as selecting a better or a worse grasp.
An object to be grasped is placed in the simulation according to a single \textbf{reference pose} for that object. 
A reference grasp for each gripper $g$ is handcrafted for each object $i$ in its reference pose.
All selected objects and their reference grasps for one gripper are shown in Fig.~\ref{fig:reference_grasps}.
We do not consider grasp optimality in this study. Reference grasps are chosen to be sensible, and better or worse reference grasps would merely shift the distributions of successes and failures.
The precise definition of grasp success used in our experiments is provided in Section \ref{sub::pose_est_grasp_metrics}.
Alternative definitions could have been adopted.

Having committed to a single reference pose per object, we align an object's true pose in the input image, $^{C_j}\mathbf{T}_{O_i}$, with that object's reference pose in the simulator, which has frame of reference $\{S\}$. That same transformation, $^{S}\mathbf{T}_{C_j}$, is then applied to the pose estimate $^{C_j}\hat{\mathbf{T}}_{O_i}^{(k)}$ made from the same input image. This transformation ensures that pose-estimate error is preserved, relative to the reference pose.
Formally, 

\begin{align}
  e_R\Big({}^{S}\mathbf{T}_{C_j}{}^{C_j}\hat{\mathbf{T}}_{O_i}^{(k)}, {}^{S}\mathbf{T}_{C_j}{}^{C_j}\mathbf{T}_{O_i}\Big) &= e_R\Big({}^{C_j}\hat{\mathbf{T}}_{O_i}^{(k)}, {}^{C_j}\mathbf{T}_{O_i}\Big) \\
  e_t\Big({}^{S}\mathbf{T}_{C_j}{}^{C_j}\hat{\mathbf{T}}_{O_i}^{(k)}, {}^{S}\mathbf{T}_{C_j}{}^{C_j}\mathbf{T}_{O_i}\Big) &= e_t\Big({}^{C_j}\hat{\mathbf{T}}_{O_i}^{(k)}, {}^{C_j}\mathbf{T}_{O_i}\Big)
\end{align}

\noindent where $e_R(\cdot, \cdot)$ and $e_t(\cdot, \cdot)$ are rotation and translation error functions, respectively. This means that the transformed estimate deviates as much from the reference pose as the image-based estimate does from the input image's ground truth. 
(Please see our video for an animated demonstration.)

This transformation is illustrated in Fig.~\ref{fig:change-of-frame}, where readers will notice that, in this case, transformation to the reference pose changes the direction in which gravity acts on the soup can. Changes to scene gravity occur rarely in the dataset. Furthermore, the datasets are static; objects may have been rearranged in between scenes, but they are never in motion.

Trajectory planning (adjusting grasps and reaches to navigate around scene clutter) is out of scope for this paper. Including scene clutter in the physics simulation would distract from our goal, which is the collection of grasping-trial outcomes for a single target object under uniform conditions. Figure~\ref{fig:change-of-frame} shows that we abstract away scene clutter in the simulator.

For every $^{C_j}\hat{\mathbf{T}}_{O_i}^{(k)}$, we run a grasping trial using two end-effectors: a parallel gripper \cite{haddadin2022franka} and an underactuated hand \cite{chen2020underactuation}. (We refer to both as ``grippers" and index the set of all grippers using $g$.) We use a simplified, rigid, and open-loop control policy to execute each grasping task. In open-loop mode, the system does not utilize any sensory feedback or make updates while the grasp is underway. Planning is performed using the initial pose estimate $^{C_j}\hat{\mathbf{T}}_{O_i}^{(k)}$.

\subsection{Grasp-Success Prediction}\label{sub::grasp_success_prediction}

We collect estimates made by $K$ recent 6-DoF pose estimators, pre-trained on the objects in the dataset we use. For each pose estimate $^{C_j}\hat{\mathbf{T}}_{O_i}^{(k)}$, for each object, in each RGB image, according to estimator $k \in K$, we now have the binary outcomes of grasping trials conducted in the physics simulator using each gripper $g$. These pose estimates and grasping trial results provide the requisite data to train and test success-prediction networks. Our networks receive the differences between pose estimates and, depending on the training configuration, the identity of the object to be grasped and the gripper to attempt the grasp.

The intuition is similar to ensembling in that $K$ pose-estimators have made predictions about the same input. However, an ensemble captures discrepancy among estimates as a measure of uncertainty \cite{lakshminarayanan2017ensembles}.
We select one of our $K$ estimators to act as the \textbf{Principal Estimator (PE)} and cast the remaining $(K - 1)$ as \textbf{Supporting Estimators (SE)}.
We then jointly predict the object pose using the PE and its uncertainty using the SEs.
The pose prediction made by the PE is the estimate according to which a grasp will be attempted.
Differences between the PE and the SEs will determine the network's prediction (the consensus) about whether that attempt will succeed (grasp success is defined in Section \ref{sub::pose_est_grasp_metrics}).
Previous work \cite{shi2021fast} uses an ensemble to quantify uncertainty, but identifying a PE as the estimate responsible for a downstream task connects uncertainty to action.

Two 6-DoF pose estimators may disagree with each other on the same object-image pair in six values: the three components of translation and three Euler angles describing rotation. Each PE-SE difference is a concatenation of signed pose-component differences, forming a vector in $\mathbb{R}^6$.
We experiment with different PEs and four training configurations. 
In its narrowest configuration, our approach trains a single MLP for every combination of object, gripper, and PE. Networks trained under the narrow configuration receive as input a vector in $\mathbb{R}^{6(K - 1)}$ made by concatenating the signed differences between the PE's pose prediction and each of the SEs' pose predictions. Training labels are the binary outcomes from our simulator trials.

In a wider configuration, networks are trained for a single gripper and all objects, in which case they require an additional input indicating which of $N$ objects is to be grasped. We denote this configuration ``MLP-O".
This makes the input for these MLPs $\mathbb{R}^{6(K - 1) + N}$, assuming a one-hot encoding of the target object.
Similarly, networks trained for a single object and all $M$ grippers need a gripper-indicator input, forming a vector in $\mathbb{R}^{6(K - 1) + M}$.
This configuration is called ``MLP-G".
Networks trained for all objects and all grippers (denoted ``MLP-OG") will naturally require the greatest number of inputs, $\mathbb{R}^{6(K - 1) + N + M}$.

\begin{figure}[!t]
\centering
  \includegraphics[width=0.9\columnwidth]{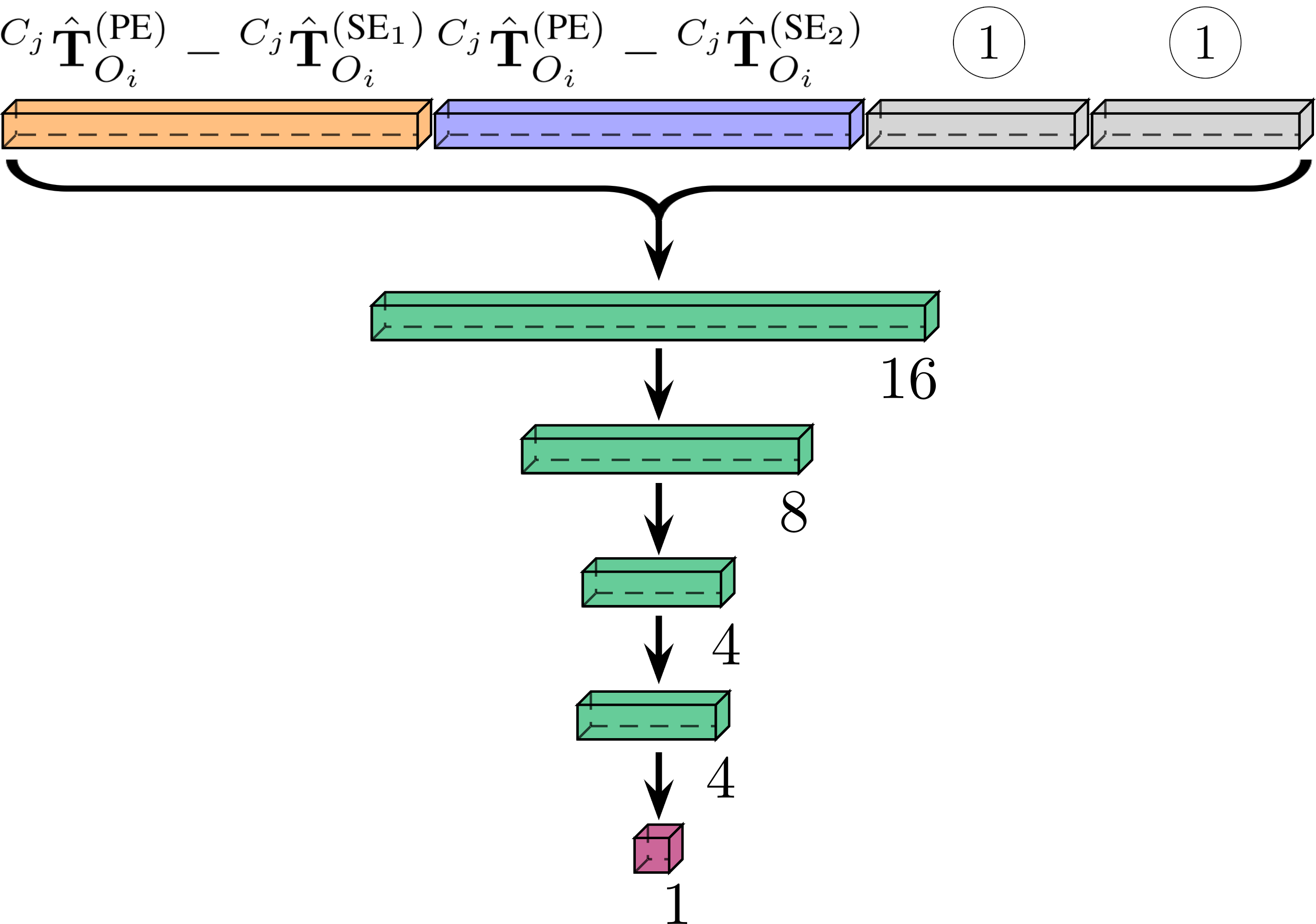}
  \caption{A diagram of our MLP architecture. All MLPs receive differences between the principal and each supporting pose estimate. For two SEs, we have two differences (orange and blue). When training on several objects, an additional one-hot vector (gray) indicates which object is to be grasped. When training on several grippers, these are identified by another one-hot vector. Green network layers use the ReLU activation function, and the output unit (pink) uses the sigmoid function.}
  \vspace{-5pt}
  \label{fig:mlp-architecture}
\end{figure}

Network architecture is depicted in Fig.~\ref{fig:mlp-architecture}. Each MLP consists of five fully-connected layers. The size of the input layer may change, according to whether objects or grippers are being specified, but the rest of the network remains the same. The first hidden layer contains 16 nodes, then subsequent layers taper down through eight, four, four, and finally one node at the output layer. All layers except the output use the ReLU activation function. The output of the MLP uses the sigmoid function, constraining values to $[0, 1]$, signifying predicted success.

Consider an MLP trained specifically for the underactuated gripper and the can of soup seen in Fig.~\ref{fig:change-of-frame}. Training and test sets for this MLP come only from grasping trials involving the soup can and underactuated gripper. If ZebraPose \cite{su2022zebrapose} is the network's PE, and EPOS \cite{hodan2020epos} and GDRNPP \cite{wang2021gdr} are SEs, then the network receives a vector of length 12, comprising two 6D differences, and outputs a scalar. The output predicts whether grasping the soup can with the underactuated gripper according to the ZebraPose estimate will succeed.

Notice that \textit{MLPs never receive the PE's pose estimate directly}. We refrain from giving the absolute pose to the network because the dataset provides insufficient coverage of pose space; the space of pose \textit{differences} is more manageable.

\section{Pose Estimation and Downstream Task Trials}\label{sec:estimate_simulate}
Here we review our implementation details and motivating observations up to but excluding network training. Introducing the range of recent estimators used in this study demonstrates that improved models do reduce pose error, but that downstream grasping is still difficult, thus necessitating uncertainty models. We describe pose estimation in detail and define outcomes for grasping trials.

\subsection{Datasets}\label{sub::datasets}
The BOP Challenge \cite{sundermeyer2023bop} unifies several datasets for training and evaluating 6-DoF pose estimators. A dataset contains one or more scenes for training and for testing. Each scene has RGB images, camera intrinsics, and ground-truth 6-DoF poses.

Our experiments use two of the more popular BOP datasets. YCB-V contains scenes of common household objects and groceries. Its RGB images have (640 $\times$ 480) resolution. We limit our experiments to seven of the least challenging items in order to obtain an upper bound for performance. Linemod-Occlusion (LM-O), also (640 $\times$ 480), contains eight objects which are present in all images and occlude each other. The LM-O objects have more complex shapes than YCB-V, being mostly small toys and handheld tools. Figure~\ref{fig:reference_grasps} shows the shapes of the objects relative to the grippers.

While 3D models of the objects included in each dataset precisely capture objects' dimensions, BOP metadata do not include information on weights or friction coefficients, which are needed for our simulations in MuJoCo \cite{mujoco}. However, objects' physical details are straightforward to estimate. We also assume that objects are non-deformable and that their densities are uniform.

\begin{table*}[!ht]
\centering
\caption{Grasping trial success rates for our three pose estimators and two grippers.}
\label{tab:grasp_success_rates}
\begin{tabular}{ll|cc|cc|cc}
\hline
    && \multicolumn{2}{c}{\textbf{EPOS} \cite{hodan2020epos}} & \multicolumn{2}{c}{\textbf{GDRNPP} \cite{wang2021gdr}} & \multicolumn{2}{c}{\textbf{ZebraPose} \cite{su2022zebrapose}} \\

    Dataset &                                                                    &
    Parallel$\uparrow$                                                           &
    Underact.$\uparrow$                                                          &
    Parallel$\uparrow$                                                           &
    Underact.$\uparrow$                                                          &
    Parallel$\uparrow$                                                           &
    Underact.$\uparrow$                                                          \\
    \hline\hline

    \textbf{YCB-V} & Cracker box &
      0.756 &
      1 &
      0.858 &
      1 &
      0.804 &
      1 \\

    \hspace{10px} & Sugar box &
      0.909 &
      0.992 &
      1 &
      1 &
      0.925 &
      1 \\

    \hspace{10px} & Soup can &
      0.344 &
      0.735 &
      0.522 &
      0.897 &
      0.599 &
      0.887 \\

    \hspace{10px} & Mustard bottle &
      0.407 &
      0.993 &
      0.907 &
      1 &
      0.547 &
      0.980 \\

    \hspace{10px} & Gelatin box &
      0.920 &
      0.933 &
      0.813 &
      1 &
      0.960 &
      1 \\

    \hspace{10px} & Potted meat can &
      0.555 &
      0.748 &
      0.724 &
      0.809 &
      0.662 &
      0.802 \\

    \hspace{10px} & Bleach cleanser &
      0.090 &
      0.723 &
      0.273 &
      0.773 &
      0.203 &
      0.763 \\ \hline

    \textbf{LM-O} & Ape &
      0.038 &
      0.339 &
      0.123 &
      0.733 &
      0.206 &
      0.789 \\

    \hspace{10px} & Can &
      0.653 &
      0.725 &
      0.849 &
      0.910 &
      0.910 &
      0.955 \\

    \hspace{10px} & Cat &
      0.053 &
      0.307 &
      0.194 &
      0.633 &
      0.314 &
      0.769 \\

    \hspace{10px} & Drill &
      0.495 &
      0.465 &
      0.565 &
      0.685 &
      0.660 &
      0.775 \\

    \hspace{10px} & Duck &
      0.273 &
      0.497 &
      0.282 &
      0.346 &
      0.578 &
      0.483 \\

    \hspace{10px} & Egg carton &
      - &
      0.189 &
      - &
      0.258 &
      - &
      0.563 \\

    \hspace{10px} & Glue &
      0.354 &
      0.510 &
      0.643 &
      0.812 &
      0.830 &
      0.911 \\

    \hspace{10px} & Hole-puncher &
      0.268 &
      0.318 &
      0.175 &
      0.190 &
      0.590 &
      0.670 \\ \hline

\end{tabular}%
\end{table*}

\begin{table*}[!ht]
\centering
\caption{Average rotation error, average translation error, and average grasp success rates per pose estimator.}
\label{tab:avg_pose_errors}
\begin{tabular}{l|cccc|cccc|cccc}
\hline
    &  
    \multicolumn{4}{c}{\textbf{EPOS} \cite{hodan2020epos}}                     &
    \multicolumn{4}{c}{\textbf{GDRNPP} \cite{wang2021gdr}}                     &
    \multicolumn{4}{c}{\textbf{ZebraPose} \cite{su2022zebrapose}}              \\

    Dataset                                                                    &
    \begin{tabular}{@{}c@{}}Rot. Err. \\ (deg)$\downarrow$\end{tabular}        &
    \begin{tabular}{@{}c@{}}Trans. Err. \\ (mm)$\downarrow$\end{tabular}       &
    \multicolumn{2}{c|}{\begin{tabular}{@{}c@{}}Success \\ 
        \begin{tabular}{@{}cc@{}} Par. $\uparrow$ & Und. $\uparrow$ \end{tabular} \end{tabular}} &

    \begin{tabular}{@{}c@{}}Rot. Err. \\ (deg)$\downarrow$\end{tabular}        &
    \begin{tabular}{@{}c@{}}Trans. Err. \\ (mm)$\downarrow$\end{tabular}       &
    \multicolumn{2}{c|}{\begin{tabular}{@{}c@{}}Success \\ 
        \begin{tabular}{@{}cc@{}} Par. $\uparrow$ & Und. $\uparrow$ \end{tabular} \end{tabular}} &
    
    \begin{tabular}{@{}c@{}}Rot. Err. \\ (deg)$\downarrow$\end{tabular}        &
    \begin{tabular}{@{}c@{}}Trans. Err. \\ (mm)$\downarrow$\end{tabular}       &
    \multicolumn{2}{c}{\begin{tabular}{@{}c@{}}Success \\ 
        \begin{tabular}{@{}cc@{}} Par. $\uparrow$ & Und. $\uparrow$ \end{tabular} \end{tabular}} \\
    
    \hline\hline

    \textbf{YCB-V} &
      10.14  &
      40.7   &
      0.569  &
      0.875  &

      6.25   &
      13.3   &
      0.728  &
      0.926  &
      
      7.12   &
      24.6   &
      0.672  &
      0.919  \\

    \textbf{LM-O} &
      23.41   &
      135.2   &
      0.305   &
      0.419   &
      
      11.05  &
      56.1   &
      0.404  &
      0.571  &
     
      9.21    &
      120.5   &
      0.584   &
      0.739   \\ \hline

\end{tabular}%
\vspace{-10pt}
\end{table*}

\subsection{Pose Estimators}\label{sub::estimators}

The estimators we have chosen form a representative set of recent works with publicly available code. We use EPOS\footnote{\url{https://github.com/thodan/epos}}, ZebraPose\footnote{\url{https://github.com/suyz526/ZebraPose}}, and GDRNPP\footnote{\url{https://github.com/shanice-l/gdrnpp_bop2022}} as provided, without any further training and without using GDRNPP's refinement module. (GDRNPP is a later iteration of GDR-Net \cite{wang2021gdr}.) In cases where authors offer several variants of their method, we use the weights that minimize rotation and translation errors on the 15 objects selected for this study.

\subsection{Grasping Success}\label{sub::pose_est_grasp_metrics}

Once the grasp is executed, and the robotic hand’s actuators close and reach equilibrium, the target object’s centroid is recorded in the hand’s base frame. After 15 seconds, during which the hand moves upward from the table, the position of the same centroid is re-evaluated relative to the hand’s base. If the relative offset between these two recorded positions exceeds tolerance, the attempt is labeled a failure. This definition covers trials in which the object is dropped and when the gripper misses the object completely, as well as significant slippages and improperly secured grasps. Otherwise, the trial is deemed a success. This allows that an object may remain in hand and shift slightly.
Our experiments use a tolerance of 5 cm.

\subsection{Grasping Trial Results}\label{sub::grasping_trial_observations}

Table~\ref{tab:grasp_success_rates} reports per-object average grasp success rates, which range from from 3.8\% to 100\%. 
Regardless of the pose estimator, both grippers perform better on YCB-V than on LM-O. Given the relatively simple, mostly-prismatic shapes of the selected YCB-V objects, this is expected behavior, which we take as validation of our simulations. Success rates for the parallel gripper lag behind those for the underactuated hand, though both rates tend to increase as the geometric errors (summarized in Table~\ref{tab:avg_pose_errors}) decrease.
Even as pose estimates improve, the concavity and curvature of the small figurines in LM-O make parallel grasps highly sensitive to error.
(Note that the dimensions of the egg carton make it impossible to define a reference grasp for our parallel gripper. Therefore, the egg carton - parallel gripper combination is excluded from experiments and statistics.)

This dramatic variance in grasp success motivates modeling uncertainty in a way that directly impacts an agent's decision to take action or to abstain. Based on these observations, we would expect to treat uncertainty for each object and gripper separately. Our findings, however, indicate that there is synergy across objects beneficial to grasp-success prediction.

\section{Success Prediction}\label{sec:success_prediction}
We can leverage the differences between pose estimates to make predictions about whether a grasp guided by the principal estimate will succeed, according to the above definition for grasp success. Here, we specify how the data from pose estimation and simulator trials are used to train MLPs. We define what makes a prediction successful, introduce our baseline method, and compare and analyze results.

\subsection{Training}\label{sub::training}

Pose estimates $^{C_j}\hat{\mathbf{T}}_{O_i}^{(k)}$ and their trial outcomes are included in our dataset if all $K = 3$ estimators make predictions for the target object and input image. We do not modify the detection results included with the estimators; if one of them fails to make an object detection in an image, we drop object $i$ in image $j$. Given the objects in this study, the estimators make predictions for 3,152 distinct object-image pairs. This yields 3,152 grasping trials for the underactuated hand. The parallel gripper has 3,010 grasping trials, since one of the objects (the egg carton) is not graspable by this gripper in any configuration. We shuffle these samples and divide them into training and test sets using an 80-20 split, ensuring that both sets have the same proportions of gripper successes and failures per object.

Each of the four network configurations was trained once with each pose estimator as the PE. Each difference between the PE's and an SE's components forms a vector in $\mathbb{R}^6$ as described in Section \ref{sub::grasp_success_prediction}. Whether learning separately (as in the row labeled MLP in Table \ref{tab:synergies}), with combined object data (MLP-O), with combined gripper data (MLP-G), or with all data (MLP-OG), networks are trained for 3,000 epochs. The Adam optimizer updates weights, and cosine annealing diminishes the learning rate from $10^{-3}$ to $10^{-5}$. Training 90 MLPs (three PEs, two grippers, 15 objects) took 32 hours on a single NVIDIA GeForce GTX 1080 Ti. We report results for the model checkpoint with best test-set accuracy.

\subsection{Baseline}\label{sub::baseline}

Shi et al. \cite{shi2021fast} have proposed a method to quantify uncertainty about a 6-DoF pose estimate using an ensemble of separately trained estimators. Their study evaluates four ``disagreement metrics" and concludes that computing the variance of the Average Distance of Distinguishable model points (ADD) \cite{hodan2020bop} among ensemble pose estimates best correlates with actual pose error between any single estimate and the ground-truth pose. ADD is defined as the average Euclidean distance between all corresponding model points when transformed by $^{C}\mathbf{T}_{O}^{(a)}$ and when transformed by $^{C}\mathbf{T}_{O}^{(b)}$. ADD is a sensible, widely used metric, but collapsing the rotation and translation error components into a single value loses information. Though deeper analysis is not included here, our grasping trials have demonstrated that translation error is a better indicator of grasp failure than rotation error. Particularly, at least 80\% of translation error occurs along the viewing direction, orthogonal to the camera's image plane. (This is to be expected, given the lack of an input depth channel.) 
Since each element of the pose affects grasping success differently, we choose to present them separately to the uncertainty predictor rather than flattening these data into a single ADD value.

Since uncertainty quantification, not robotic grasping, is the goal of Shi et al., our baseline modifies their method by applying a learned threshold to the average of ADD differences between PE and SEs. Below this threshold, the baseline predicts grasp success. To ensure fairness, we use the same training set used by the MLPs. Each of the pose estimators is cast as the baseline's PE, and we learn the optimal threshold for each tuple of (PE, object, gripper).

\begin{table*}[!ht]
\centering
\caption{Average prediction success over all objects.}
\label{tab:synergies}
\resizebox{\textwidth}{!}{%
\begin{tabular}{lcc|cc|cc|cc|cc}
\hline
    & All & All & \multicolumn{2}{c}{\textbf{EPOS} \cite{hodan2020epos}} & \multicolumn{2}{c}{\textbf{GDRNPP} \cite{wang2021gdr}} & \multicolumn{2}{c}{\textbf{ZebraPose} \cite{su2022zebrapose}} & \multicolumn{2}{c}{\textbf{Avg.}} \\

    Method                                                                       &
    Objects                                                                      &
    Grippers                                                                     &
    Parallel$\uparrow$                                                           &
    Underact.$\uparrow$                                                          &
    Parallel$\uparrow$                                                           &
    Underact.$\uparrow$                                                          &
    Parallel$\uparrow$                                                           &
    Underact.$\uparrow$                                                          &
    Parallel$\uparrow$                                                           &
    Underact.$\uparrow$                                                          \\
    \hline\hline

    Baseline \cite{shi2021fast} & & &
      0.792 &
      0.863 &
      0.840 &
      0.849 &
      0.759 &
      0.863 &
      0.797 &
      0.858 \\

    MLP & & &
      0.844 &
      0.865 &
      0.869 &
      0.916 &
      0.849 &
      0.893 &
      0.854 &
      0.891 \\

    MLP-O & \checkmark & &
      \textbf{0.898} &
      \textbf{0.905} &
      0.890 &
      \textbf{0.926} &
      \textbf{0.875} &
      0.909 &
      \textbf{0.888} &
      \textbf{0.913} \\

    MLP-G & & \checkmark &
      0.872 &
      0.895 &
      \textbf{0.892} &
      0.921 &
      0.849 &
      0.903 &
      0.871 &
      0.906 \\

    MLP-OG & \checkmark & \checkmark &
      0.877 &
      0.892 &
      0.879 &
      0.907 &
      0.871 &
      \textbf{0.912} &
      0.876 &
      0.904 \\ \hline

\end{tabular}%
}
\vspace{-10pt}
\end{table*}

\subsection{Quantitative Results}\label{sub::quant}

A method should predict grasp success or failure when the PE succeeds or fails in the given trial, respectively.
We evaluate both the baseline and our MLPs on this basis.

As seen in the row of Table \ref{tab:synergies} labeled ``MLP", on average, MLPs trained for specific tuples of (PE, object, gripper) outperform the baseline for all PEs and grippers.
Combining all object-data and training MLPs for each PE and gripper improves per-object prediction further, as evidenced in the row labeled ``MLP-O". The MLP-O variant provides the MLPs with additional input to identify the target object but preserves the network architecture. The general superiority of the jointly-trained networks MLP-O suggests that, though object grasp success rates can vary drastically (for the GDRNPP PE and the parallel gripper in Table \ref{tab:grasp_success_rates}, for instance, where success rates range from 12\% to 85\%), there is still synergy across objects beneficial to the networks.

This is less true across grippers. For the row in Table \ref{tab:synergies} labeled ``MLP-G", networks are trained per object using data from both parallel and underactuated trials. MLP input is expanded to identify which gripper will attempt the grasp. Performance for this configuration improves over the baseline and separate training, though less resoundingly. This shows that the grippers used in this study are too dissimilar to facilitate learning. Interestingly, training MLPs using the entire training set (``MLP-OG") is less beneficial than using all objects. In the MLP-OG configuration, networks receive both the identities of the object to be grasped and the gripper to attempt the grasp. The MLP-OG networks either become confused by the variety of data or strain to absorb a greater number of inputs.

\subsection{Discussion}\label{sub::discussion}

Without a PE, the baseline method quantifies uncertainty but does not know how to grasp, and without a learned threshold, it provides no indication of whether to attempt or refuse a grasp. We have demonstrated that 6-DoF pose predictions can be used both to guide an open-loop grasp and to provide an uncertainty estimate that strongly correlates with the grasp's probability for success.

Our studies of pose estimator performance and grasp failure rates indicate that rotation and translation affect the outcomes of downstream grasps differently. 
We attribute the improved performance of our success-prediction method over the baseline to the preservation of geometric subtleties in $\mathbb{R}^6$ (lost in ADD) and to the MLP's ability to learn these patterns.

It is tempting to augment MLP-training data by perturbing pose estimates in the physics simulator. This is partially the approach taken by Naik et al. \cite{naik2024robotic} as they use perturbed poses to explore ranges of tolerable error in their simulated grasping task. Naik et al. use these bounds as probabilistic bins for predicting task failure. This binning respects the observation that uncertainty about pose is not Gaussian.
Likewise, Haugaard et al. \cite{haugaard2023spyropose} remark that pose uncertainty, as determined by object symmetries and occlusions, does not behave like a normal distribution. We agree with these comments; \textit{poses perturbed by Gaussian noise will not fail the same way that real pose estimators fail on real images.} The elements of the 6D difference vectors are correlated with each other in practice. For example, translation errors compensate for rotation errors. Generating grasping trials in the simulator by perturbing pose estimates would better cover the space of pose-estimate differences, but these samples would not have the true covariance structure over the elements of the poses and errors.

\section{Conclusions}\label{sec:conclusions}
As urged by \cite{thalhammer2024challenges}, uncertainty for 6-DoF pose estimation demands further study, though it can be a difficult subject to approach. We have proposed a method that models uncertainty and connects it to application. Examining how image-based object pose estimation impacts the downstream task of robotic grasping led us to observe that the geometric errors that cause task failure are too subtle to be collapsed into a single value. We therefore designed a simple but effective model to predict task failure before the task is attempted. On their own, these networks outperform the baseline by an average of 4.5\%. Improvement jumps to 7.27\% when networks are trained on all objects. Despite their variable performance in grasping trials, enough synergy exists across objects to inform a more general grasping task. We do not observe the same gains when the training process groups together all grippers or all objects and all grippers. We attribute this to the large gap between our two grippers.

{\small
\bibliographystyle{IEEEtran}
\bibliography{bib/bibliography, bib/hand_designs}
}

\vfill

\end{document}